\documentclass{article} 

\usepackage{arxiv}

\title{Towards Feature Space Adversarial Attack}

\usepackage{amsmath,amsfonts,bm}

\def\eqref#1{equation~\ref{#1}}

\def\1{\bm{1}}

\def\vm{{\bm{m}}}

\def\vx{{\bm{x}}}

\DeclareMathAlphabet{\mathsfit}{\encodingdefault}{\sfdefault}{m}{sl}
\SetMathAlphabet{\mathsfit}{bold}{\encodingdefault}{\sfdefault}{bx}{n}

\def\sA{{\mathbb{A}}}

\def\sS{{\mathbb{S}}}

\newcommand{\E}{\mathbb{E}}

\newcommand{\R}{\mathbb{R}}

\DeclareMathOperator*{\argmax}{arg\,max}
\DeclareMathOperator*{\argmin}{arg\,min}

\newcommand{\todoc}[2]{{\textcolor{#1}{\textbf{#2}}}}
\newcommand{\todored}[1]{{\todoc{red}{\textbf{[#1]}}}}

\newcommand{\todoblue}[1]{\todoc{blue}{\textbf{[#1]}}}
\newcommand{\todoorange}[1]{\todoc{orange}{\textbf{[#1]}}}

\newcommand{\xz}[1]{\todored{XZ: #1}}
\newcommand{\gt}[1]{\todoblue{GT: #1}}
\newcommand{\ql}[1]{\todoorange{QL: #1}}

\usepackage{microtype}
\usepackage{graphicx}
\usepackage{caption}
\usepackage{booktabs} 
\usepackage{subcaption}
\usepackage{siunitx}
\usepackage[hyphenbreaks]{breakurl}
\usepackage{multirow}
\usepackage{color}
\usepackage{xcolor}
\usepackage[colorlinks = true,
            linkcolor = purple,
            urlcolor  = blue,
            citecolor = cyan,
            anchorcolor = black]{hyperref}	
\usepackage[numbers,square]{natbib}
\usepackage{float}

\usepackage[utf8]{inputenc} 
\usepackage[T1]{fontenc}    
\usepackage{url}            
\usepackage{nicefrac}       
\usepackage{microtype}      

\usepackage{amsmath}
\newcommand{\defeq}{\overset{\text{\tiny def}}{=}}

\usepackage{hyperref}

\author{
 Qiuling Xu \\
  Department of Computer Science\\
  Purdue Univeristy\\
  \texttt{xu1230@purdue.edu} \\
   \And
 Guanhong Tao \\
  Department of Computer Science\\
  Purdue Univeristy\\
  \texttt{taog@purdue.edu} \\
  \And
 Siyuan Cheng \\
  Department of Computer Science\\  Purdue Univeristy\\
  \texttt{cheng535@purdue.edu} \\
  \And
 Xiangyu Zhang \\
  Department of Computer Science\\ Purdue Univeristy\\
  \texttt{xyzhang@cs.purdue.edu} \\
}

\begin{document}

\twocolumn[
\maketitle
\begin{abstract}
\noindent We propose a new adversarial attack to Deep Neural Networks for image classification. Different from most existing attacks that directly perturb input pixels, our attack focuses on perturbing abstract features, more specifically, features that denote styles, including interpretable styles such as vivid colors and sharp outlines, and uninterpretable ones. It induces model misclassfication by injecting imperceptible style changes 
through an optimization procedure.
We show that our attack can generate adversarial samples that are more natural-looking than the state-of-the-art unbounded attacks. The experiment also supports that existing pixel-space adversarial attack detection and defense techniques can hardly ensure robustness in the style related feature space. \footnotemark 
\end{abstract}
\vspace{0.35cm}
]
\footnotetext{The code is available at  \burl{https://github.com/qiulingxu/FeatureSpaceAttack}.}

\section{Introduction}

Adversarial attacks are a prominent threat to the broad application of Deep Neural Networks (DNNs). In the context of classification applications, given a pre-trained model $M$ and a benign input $\vx$ of some output label $y$, adversarial attack perturbs $\vx$ such that $M$ misclassifies the perturbed $\vx$. The perturbed input is called an {\em adversarial example}.  Such perturbations are usually bounded by some distance norm such that they are not perceptible by humans. Since it was proposed in~\citep{szegedy2013intriguing}, there has been a large body of research that develops various methods to construct adversarial examples
with different modalities (e.g., images~\citep{carlini2017towards}, audio~\citep{qin2019imperceptible}, text~\citep{ebrahimi2018hotflip}, and video~\citep{li2018adversarial}), to detect adversarial examples~\citep{tao2018attacks,ma2019nic}, and use adversarial examples to harden models~\citep{madry2017towards,zhang2019theoretically}. 

\begin{figure}[!ht]
    \centering
    \begin{subfigure}[t]{.11\textwidth}
        \centering\includegraphics[width=\textwidth]{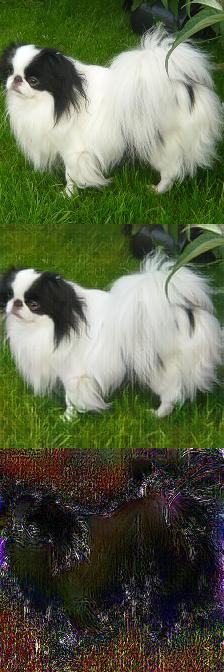}
        \caption{Spaniel\\$\ell_\infty$:121/255\\$\ell_2$:25.92}
        \label{fig:my_label}
    \end{subfigure}
    \begin{subfigure}[t]{.11\textwidth}
        \centering\includegraphics[width=\textwidth]{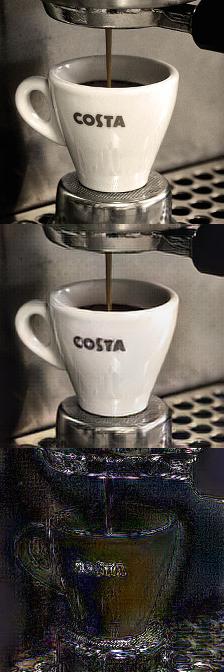}
        \captionsetup{justification=centering}
        \caption{Espresso\\192/255\\24.47}
        \label{fig:my_label}
    \end{subfigure}
    \begin{subfigure}[t]{.11\textwidth}
        \centering\includegraphics[width=\textwidth]{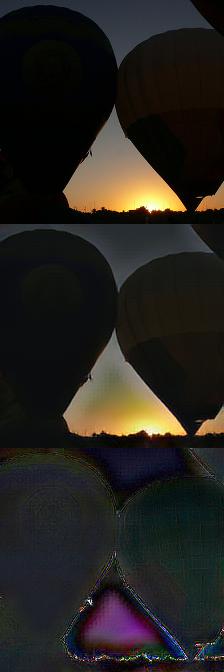}
        \captionsetup{justification=centering}
        \caption{Balloon\\149/255\\20.75}
        \label{fig:my_label}
    \end{subfigure}
    \begin{subfigure}[t]{.11\textwidth}
        \centering\includegraphics[width=\textwidth]{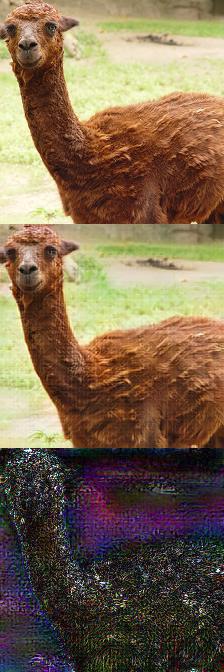}
        \captionsetup{justification=centering}
        \caption{Llama\\183/255\\28.55}
        \label{fig:my_label}
    \end{subfigure}
    \begin{subfigure}[t]{.11\textwidth}
        \centering\includegraphics[width=\textwidth]{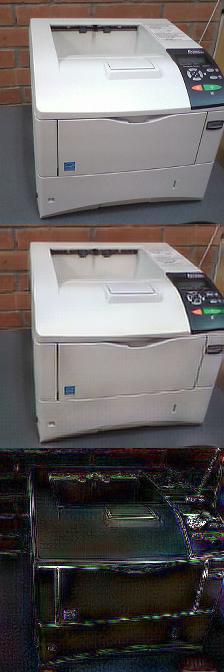}
        \captionsetup{justification=centering}
        \caption{Printer\\252/255\\21.80}
        \label{fig:my_label}
    \end{subfigure}
    \begin{subfigure}[t]{.11\textwidth}
        \centering\includegraphics[width=\textwidth]{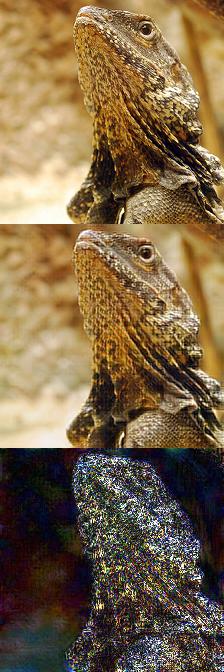}
        \captionsetup{justification=centering}
        \caption{Lizard\\225/255\\40.88}
        \label{fig:my_label}
    \end{subfigure}
     \begin{subfigure}[t]{.11\textwidth}
        \centering\includegraphics[width=\textwidth]{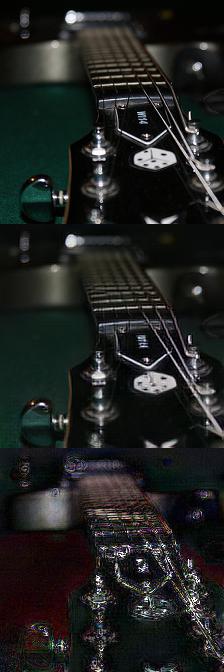}
        \captionsetup{justification=centering}
        \caption{Guitar\\216/255\\25.60}
        \label{fig:my_label}
    \end{subfigure}
    \begin{subfigure}[t]{.11\textwidth}
        \centering\includegraphics[width=\textwidth]{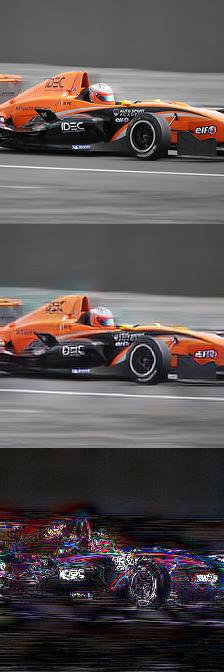}
        \captionsetup{justification=centering}
        \caption{Race car\\248/255\\29.67}
        \label{fig:my_label}
    \end{subfigure}
    \caption{Examples by feature space attack. The top row presents the original images. The middle row denotes the adversarial samples. The third row shows the pixel-wise difference ($\times 3$) between the original and the adversarial samples. The $\ell_\infty$ and $\ell_2$ norms are shown on the bottom.
    }
    \label{fig:attack_examples}
\end{figure}

However, most existing attacks (in the context of image classification) are in the 
pixel space. That is, bounded perturbations are directly applied to pixels.
In this 
paper, we illustrate that adversarial attack 
can be conducted in the style related feature space. The underlying assumption is that during training, a DNN may extract a large number of abstract features. While many of them denote critical characteristics of the object, some of them are secondary, for example, the different styles of an image (e.g., vivid colors versus pale colors, sharp outlines versus blur outlines). 
These secondary features may play an improperly important role in model prediction. As a result, feature space attack can inject such 
secondary features, which are not simple pixel perturbation, but rather functions over the given benign input, to induce model misclassification. Since humans are not sensitive to these features, the resulted adversarial examples look  natural from humans' perspective. As many of these features
are pervasive, the resulted pixel space perturbation may be much more  substantial than existing pixel space attacks. As such, pixel space defense techniques may become ineffective for 
feature space attacks (Section~\ref{sec:eval}).
Figure~\ref{fig:attack_examples} shows a number of adversarial examples generated by our technique, their comparison with the original examples, and the pixel space distances.
Observe that while the distances are much larger compared to those in pixel space attacks, the adversarial examples are  natural, or even indistinguishable from the original inputs in humans' eyes. The contrast of the benign-adversarial pairs illustrates that the malicious perturbations largely co-locate with the primary content features, denoting imperceptible style changes.

Under the hood, we consider that the activations of  an inner layer represent a set of abstract features, including those primary and secondary. 
Distinguishing the two types of features is crucial for the quality of feature-space attack.  To avoid generating
adversarial examples that are unnatural, we refrain from tampering with the primary features (or {\em content features}) and focus on perturbing the secondary {\em style features}. 
Inspired by the recent advance in style transfer~\citep{huang2017arbitrary}, the {\em mean} and
{\em variance} of activations are considered the style. As such, we focus on perturbing the
means and variances while preserving the {\em shape} of the activation values (i.e., the up-and-downs of these values and the relative scale of such up-and-downs). We use gradient driven optimization to search for the style perturbations that can induce misclassification. Since our threat model is the same as existing pixel space attacks, that is, the attack is launched by providing the adversarial example to the model. An important step is to translate the activations with style changes back to a naturally looking pixel space example. We address the problem by considering the differences of any pair of training inputs of the same class as the possible style differences, and pre-training a decoder that can automatically impose styles in the pixel space based on the style feature variation happening in an inner layer.
We propose two concrete feature space attacks, one to enhance styles and the other to impose styles constituted from a set of pre-defined style prototypes. 

We evaluate our attacks on 3 datasets and 7 models. We show  that feature space attacks can effectively generate adversarial samples.
The generated samples have natural, and in many cases, human imperceptible style differences compared with the original inputs. Our comparative experiment with  recent attacks on colors~\cite{hosseini2018semantic} and semantics~\cite{semantic_attack} shows that our generated samples are more natural-looking.
We also show that 7 state-of-the-art detection/defense approaches are ineffective to our attack as they focus on protecting the pixel space.  Particularly, our attack reduces the detection rate of a state-of-the-art pixel-space  approach~\citep{roth2019odds} to 0.04\% on the CIFAR-10 dataset, and the prediction accuracy of a model hardened by a state-of-art pixel-space adversarial training technique \citep{xie2019feature} to 1.25\% on ImageNet. Moreover, we observe that despite the large distance introduced in the pixel space (by our attack), the distances in feature space are similar or even smaller than those in $\ell$-norm based attacks.
Note that the intention of these experiments is not to claim our attack is superior, but rather to illustrate that new defense and hardening techniques are needed for feature space protection.


\section{Background and Related Work}
\label{sec:background}


\noindent
{\bf Style Transfer.}
~\citet{huang2017arbitrary} proposed to transfer the style from a (source) image to another (target) that may have different content such that the content of the target image largely retains while features that are not essential to the content align with those of the source image.
Specifically, given an input image, say the portrait of actor Brad Pitt, and a style picture, e.g., a drawing of painter Vincent van Gogh, the goal of style transfer is to produce a portrait of Brad Pitt that looks like a picture painted by Vincent van Gogh. Existing approaches leverage various techniques to achieve this purpose. \citet{gatys2016image} utilized the feature representations in convolutional layers of a DNN to extract {\em content features} and {\em style features} of input images. Given a random white noise image, the algorithm feeds the image to the DNN to obtain the corresponding content and style features. The content features from the white noise image are compared with those from a content image, and the style features are contrasted with those from a style image. It then minimizes the above two differences to transform the noise image to a content image with style. 
Due to the inefficiency of this optimization process, researchers replace it with a neural network that is trained to minimize the same objective \citep{li2016precomputed,johnson2016perceptual}. Further study extends these approaches to synthesize more than just one fixed style \citep{dumoulin2016learned,li2017diversified}. 
\citet{huang2017arbitrary} introduced a simple and yet effective approach, which can efficiently enable arbitrary style transfer. It proposed an {\em adaptive instance normalization} (AdaIN) layer that aligns the mean and variance of the content features with those of the style features. \\
\noindent
{\bf Adversarial Attacks beyond Pixel Space.}
The exploration beyond $\ell$-norm based attacks is rising. \citet{transfer} found that simulating feature representation of target label improves transferability.  \citet{hosseini2018semantic} proposed to modify the HSV color space to generate adversarial samples. The method transforms all pixels by a non-parametric function uniformly. Differently, our feature space attack changes colors of objects or background and the transformation is learned from images of the same object with different styles. It is hence more imperceptible.
\citet{laidlaw2019functional} proposed to change the lighting condition and color (like \cite{hosseini2018semantic}) to generate adversarial examples. 
\citet{artattack} produced art-style images as adversarial samples. 
It does not restrict the feature space 
such that the generated samples are not natural looking, especially compared to ours. 
\citet{semantic_attack} generated semantic adversarial examples by modifying color and texture.  
It advocates not to restrict  attack space and is hence considered unbounded. 
As such, it is difficult to control the attack to avoid generating unrealistic samples. 
In contrast, our attack has a well-defined attack space while being unbounded in the pixel space. 
It implicitly learns to modify lighting condition, color and texture, it tends to be more general and capable of transforming subtle (and uninterpretable) features (Section \ref{sec:human_study}). 
Unlike in \citet{song2018constructing}, where a vanilla GAN-based attack generates samples over a distribution of limited support, and has little control of the generated samples, our encoder-decoder based structure enables attacking individual samples with controlled content. 
\citet{disentangle} proposed to perturb the latent embedding of VAE-GAN to generate adversarial samples. Since it does not distinguish primary and secondary features, the perturbation on primary features substantially degrades the quality of generated adversarial samples. In contrast, our feature space perturbation is more effective. 

We empirically compare with these attacks later in the paper.

\section{Feature Space Attack}
\begin{figure}[t]
    \centering
    \begin{subfigure}[t]{.49\textwidth}
        \centering\includegraphics[width=\textwidth]{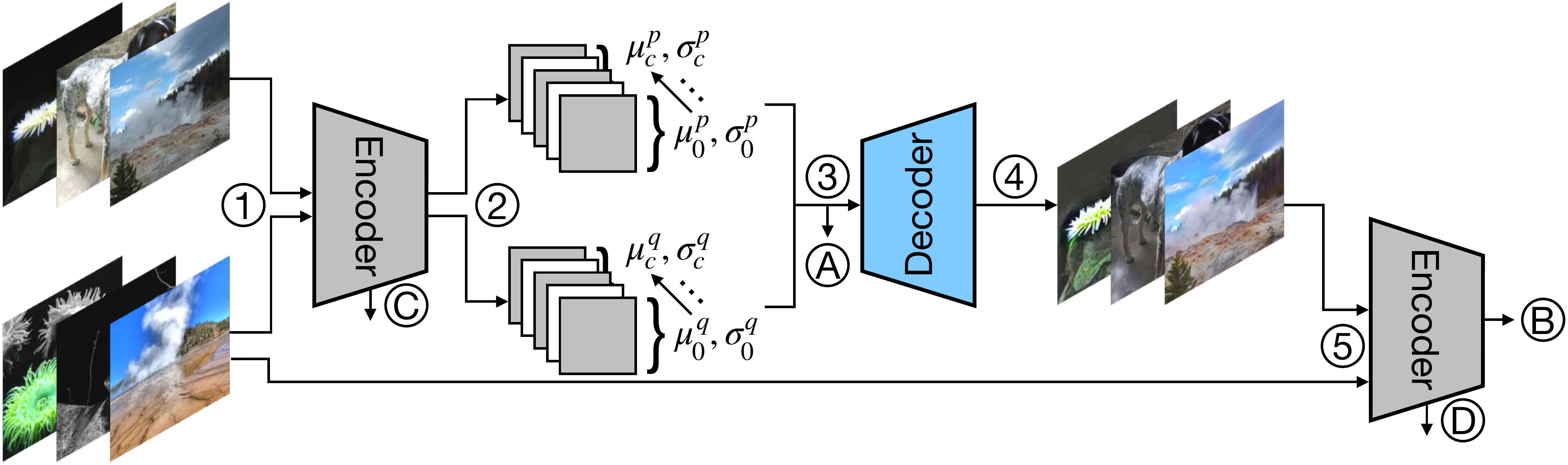}
        \vspace{-20pt}
        \caption{Decoder training phase }
        \label{fig:decoder}
    \end{subfigure}
    \begin{subfigure}[t]{.49\textwidth}
        \centering\includegraphics[width=\textwidth]{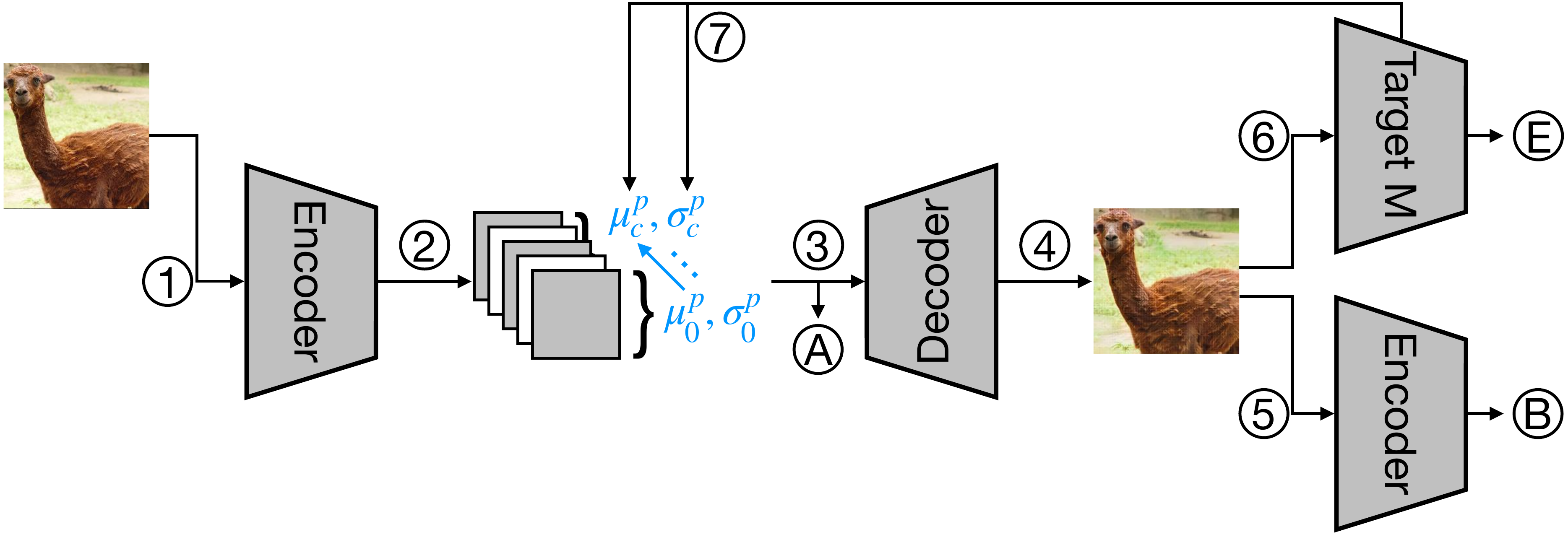}
        \vspace{-20pt}
        \caption{Feature space attack }
        \label{fig:attack}
    \end{subfigure}
    \vspace{-5pt}
    \caption{Procedure of feature space adversarial attack. Two phases are involved during the attack generation process: (a) decoder training phase and (b) feature space attack phase.}
    \label{fig:architecture}
\end{figure}

{\bf Overview.} 
We aim to demonstrate that perturbation in the feature space can lead 
to model misbehavior, which existing pixel space defense techniques cannot effectively defend against. The hypothesis is that during training, the model picks up numerous features, many of which do not describe the key characteristics (or {\em content}) of the object, but rather human imperceptible features such as styles. These subtle features may play an improperly important role in model prediction. As a result, injecting such features to a benign image can lead to misclassification. However, the feature space is not exposed to attackers such that they cannot directly perturb features.
Therefore, a prominent challenge is to derive the corresponding pixel space mutation that appears natural to humans while leading to the intended feature space perturbation, and eventually the misclassification. 
In particular, the attack comprises two phases: (1) training a decoder that can translate feature space
perturbation to pixel level changes that look natural for humans;
(2) launching the attack by first using gradient based optimization to identify feature space perturbation that can cause misclassification and then using the decoder to generate the corresponding adversarial example.
Inspired by style transfer techniques, we consider a much
confined feature perturbation space -- {\em style perturbation}. Specifically, 
as in ~\citet{huang2017arbitrary}, we consider the {\em mean} and {\em variance} of the activations
of an inner layer denote the style of the features in that layer whereas the activations themselves denote the content features. We hence perturb the mean and
variance of content features by performing a predefined transformation that largely preserves the shape of the features while changing the mean and variance. The decoder then decodes the perturbed feature values to an image closely resembles the original image with only style differences that appear natural to humans but causing model misclassification.


Fig.~\ref{fig:architecture} illustrates the workflow of the proposed attack. In the Decoder training phase (a), a set of image pairs with each pair from the same class (and hence their differences can be intuitively considered as style differences) are fed to a fixed Encoder that essentially consists of the first a few layers of a pre-trained model (e.g., VGG-19) (step \textcircled{1}). The Encoder produces the internal embeddings of the two respective images, which correspond to the activation values of some inner layer in the pre-trained model, e.g., conv4\_1 (step \textcircled{2}). Each internal embedding consists of a number of matrices, one for every channel.
For each embedding matrix, the mean and variance are computed. We use these values from the two input images to produce the integrated embedding \textcircled{A} (step \textcircled{3}), which will be discussed in details later in this section. Intuitively, it is generated by performing a shape-preserving transformation of the upper matrix so that it retains the content features denoted by the upper matrix while having the
mean and variance of the lower matrix (i.e., the style denoted by the lower matrix).
We employ a Decoder to reconstruct a raw image from \textcircled{A} at step \textcircled{4}, which is supposed to have the content of the upper image (called the {\em content image}) and the style of the lower image (called the {\em style image}). To enable good reconstruction performance, two losses are utilized for optimizing the Decoder. The first one is the {\em content loss}. Specifically, at step \textcircled{5} the reconstructed image is passed to the Encoder to acquire the reconstructed embedding \textcircled{B}, and then the difference between the integrated embedding \textcircled{A} and the reconstructed embedding \textcircled{B} is minimized. The second one is the {\em style loss}. Particularly, the means and variances of a few selected internal layers of the Encoder are computed for both the generated image and the original style image. The difference of these values of the two images is minimized. The Decoder optimization process is conducted on the original training dataset of target model $M$ (under attack).
Intuitively, the decoder is trained to understand the style differences so that it can decode feature style differences to realistic pixel space style differences, by observing the possible style differences.

When launching the attack ((b) in Fig.~\ref{fig:architecture}), a test input image is fed to the Encoder and goes through the same process as in the Decoder training phase. The key differences are that only one input image is required and the Decoder is fixed in this phase. Given a target model $M$ (under attack), the reconstructed image is fed to $M$ at step \textcircled{6} to yield prediction \textcircled{E}. As the attack goal is to induce $M$ to misclassify, the difference between prediction \textcircled{E} and a target output label (different from \textcircled{E}) is considered the {\em adversarial loss} for launching the attack. In addition, the content loss between \textcircled{A} and \textcircled{B} is also included. The attack updates the means and variances of embedding matrices at step \textcircled{7} with respect to the adversarial loss and content loss.
The final reconstructed image that induces the target model $M$ to misclassify is a successful adversarial sample.

\subsection{Definitions}
In this section, we formally define feature space attack. Considering a typical classification problem, where the samples $\vx \in \R^d$ and the corresponding label $y \in \{0, 1, \dots, n \}$ jointly obey a distribution $\mathcal{D}(\vx,y)$. Given a classifier $M: \R^d \rightarrow \{0, 1, \dots, n \}$ with parameter $\theta$. The goal of training is to find the best parameter $\argmax_\theta P_{(\vx,y) \sim \mathcal{D}}[M(\vx;\theta)=y]$. Empirically, people associate a continuous loss function $\mathcal{L}_{M,\theta}(\vx,y)$, e.g. cross-entropy, to measure the difference between the prediction and the true label. 
And the goal is rewritten as $\argmin_\theta \E_{(\vx,y) \sim \mathcal{D}}[\mathcal{L}_{M,\theta}(\vx, y)]$.
We use $\mathcal{L}_{M}$ in short for $\mathcal{L}_{M,\theta}$ in the following discussion.
In adversarial learning, the adversary can introduce a perturbation $\boldsymbol\delta \in \sS \subset \R^d$ to a natural samples $(\vx, y) \sim \mathcal{D}$. For a given sample $\vx$ with label $y$, an adversary chooses the most malicious perturbation $\argmax_{\boldsymbol\delta \in \sS}{\mathcal{L}_{M}(\vx+\boldsymbol\delta,y)}$ to make the classifier $M$ predict incorrectly. 
Normally $\sS$ is confined as an $\ell_p$-ball centered on $0$. In this case, the $\ell_p$ norm of pixel space differences measures the distance between adversarial samples (i.e., $\vx+\boldsymbol\delta$ that causes misclassification) and the original samples. Thus we refer to this attack model as the {\em pixel space attack}.
Most existing adversarial attacks fall into this category.
Different from adding bounded perturbation in the pixel space, feature space attack applies perturbation in the feature space such that an encoder (to extract the feature representation of the benign input) and a decoder function (that translates perturbed feature values to a natural-looking image that closely resembles the original input in humans' perspective).

Formally, consider an encoder function $f: \R^d \rightarrow \R^{e}$ and a decoder function $f^{-1}: \R^{e} \rightarrow \R^{d}$. The former encodes a sample to an embedding $b \in \R^{e}$ and the latter restores an embedding back to a sample. A perturbation function $a \in \sA : \R^e \rightarrow \R^{e}$ transforms a given embedding to another.
For a given sample $\vx$, the adversary chooses the best perturbation function to make the model $M$ predict incorrectly.

\begin{equation}
\max_{a \in \sA}\mathcal{L}_{M}[ f^{-1} \circ a \circ f (\vx),y].
\label{eq:goal}
\end{equation}
Functions $f$ and $f^{-1}$ need to satisfy additional properties to ensure the attack is meaningful. We call them the {\em wellness properties} of encoder and decoder.

\noindent
{\em Wellness of Encoder $f$.} In order to get a meaningful embedding, there ought to exist a well-functioning classifier $g$ based on the embedding, with a prediction error rate less than $\delta_1$.

\begin{equation}
\begin{split}
     \exists g: \R^e \rightarrow  \{0, 1, \dots, n \}, &  P_{(\vx,y) \sim \mathcal{D}}[g(f(\vx)) = y] \\
      \geq 1 - \delta_1, \;\mathrm{for\;a\;given}\; \delta_1 &.    
\end{split}
\end{equation}
In practice, this property can be easily satisfied as one can construct $g$ from a well-functioning classifier $M$, by decomposing $M = M_2 \circ M_1$ and take $M_1$ as $f$ and $M_2$ as $g$.

\noindent
{\em Wellness of Decoder $f^{-1}$.} Function $f^{-1}$ is essentially a translator that translates what the adversary has done on the embedding back to a sample in $\R^d$. We hence require that
for all possible adversarial transformation $a \in \sA$, $f^{-1}$ ought to retain what the adversary has applied to the embedding in the restored sample. 
\begin{equation}
    \begin{split}
     \forall a\in \sA, \text{ let } B^a = a \circ f(\vx), \; 
     E_{(\vx,y) \sim \mathcal{D}} \\ ||f \circ f^{-1} (B^a) - B^a||_2 
     \leq \delta_2 ,\mathrm{\;for\;a\;given\;\delta_2}.
    \end{split}
    \label{reconstruct_loss}
\end{equation}
This ensures a decoded (adversarial) sample induce the intended perturbation in the feature space.
Note that $f^{-1}$ can always restore a benign sample back to itself. This is equivalent to requiring the identity function in the perturbation function set $\sA$.  
    
Given $(f,f^{-1},\sA)$ satisfying the aforementioned properties, we define Eq.~(\ref{eq:goal}) as a feature space attack. 
Under this definition, pixel space attack is a special case of feature space attack. For an $\ell_p$-norm $\epsilon$-bounded pixel space attack, i.e., $\sS= \{||\boldsymbol\delta||_p \leq \epsilon\}$, we can rewrite it as a feature-space attack. Let encoder $f$ and decoder $f^{-1}$ be an identity function and let $\sA = \cup_{||\boldsymbol\delta||_p \leq \epsilon} \{a : a(\vm) = \vm + \boldsymbol\delta \}$. 

One can easily verify the wellness of $f$ and $f^{-1}$.
Note that the stealthiness of feature space attack depends on the selection
of $\sA$, analogous to that the stealthiness of pixel space attack depending on the $\ell_p$ norm.
Next, we demonstrate two stealthy feature space attacks.

\subsection{Attack Design}

\noindent
{\bf Decoder Training.}
Our decoder design is illustrated in Fig.~\ref{fig:decoder}. It is inspired by
style transfer in \citep{huang2017arbitrary}.
To train the decoder, we enumerate all the possible pairs of images in each class in the original training set and use these pairs as a new training set.
We consider each pair has the same content features (as they belong to the same class) and hence their differences essentially denote style differences. 
By training the decoder on all possible style differences (in the training set) regardless the output classes, we have a general decoder that can recognize and
translate arbitrary style perturbation.
Formally, given a normal image $\vx_p$ and another image $\vx_q$ from the same class as $\vx_p$, the training process first passes them through a pre-trained Encoder $f$ (e.g., VGG-19) to obtain embeddings $B^p = f(\vx_p), B^q = f(\vx_q) \in \R^{H \cdot W \cdot C}$, where $C$ is the channel size, and $H$ and $W$ are the height and width of each channel. For each channel $c$, the mean and variance are computed across the spatial dimensions (step \textcircled{2} in Fig.~\ref{fig:decoder}). That is,
\begin{equation}
\begin{split}
     & \mu_{B_c} = \frac{1}{HW} \sum_{h=1}^H \sum_{w=1}^W B_{hwc}\, \\
     & \sigma_{B_c} = \sqrt{\frac{1}{HW} \sum_{h=1}^H \sum_{w=1}^W (B_{hwc} - \mu_{B_c})^2}\,.
\end{split}
\label{eq:mean_var}
\end{equation}
We combine the embeddings $B^p$, $B^q$ from the two input images using the following equation:
\begin{equation}
    \forall c\in [1,2,...,C], B_c^o = \sigma_{B_c^q} \Big ( \frac{B_c^p - \mu_{B_c^p}}{\sigma_{B_c^p}} \Big) + \mu_{B_c^q},
\end{equation}
where $B_c^o$ is the result embedding of channel $c$. Intuitively, the transformation retains the shape of $B^p$ while enforcing the mean and 
variance of $B^q$. $B^o$ is then fed to the Decoder $f^{-1}$ for reconstructing the image with the content of $\vx_p$ and the style of $\vx_q$ (steps \textcircled{3} \& \textcircled{4} in Fig.~\ref{fig:decoder}). In order to generate a realistic image, the reconstructed image is passed to Encoder $f$ to acquire the reconstructed embedding $B^r=f \circ f^{-1}(B^o)$ (step \textcircled{5}). The difference between the combined embedding $B^o$ and the reconstructed embedding $B^r$, called the {\em content loss}, is minimized 
using the following equation during the Decoder training:
\begin{equation}
\label{eq:content_loss}
    \mathcal{L}_{\mathrm{content}} = ||B^r - B^o||_2.
\end{equation}
Note that the similarity between the input and the output is implicitly ensured by the fact that the encoder is relatively shallow and well-trained.
In addition, some internal layers of Encoder $f$ are selected, whose means and variances (computed by Equation~\ref{eq:mean_var}) are used for representing the style of input images. The difference of these values between the style image $\vx_q$ and the reconstructed image $\vx_r$, called the {\em style loss}, is minimized when training the Decoder. It is defined as follows:
\begin{equation}
\begin{split}
    & \mathcal{L}_{\mathrm{style}}  =   \sum_{i\in L} ||\mu(\phi_i(\vx_q)) - \mu(\phi_i(\vx_r))||_2   \\ + 
      & \sum_{i\in L} ||\sigma(\phi_i(\vx_q)) - \sigma(\phi_i(\vx_r))||_2
\end{split}
\end{equation}
where $\phi_i(\cdot)$ denotes layer $i$ of Encoder $f$ and $L$ the set of layers considered. In this paper, $L$ consists of conv1\_1, conv2\_1, conv3\_1 and conv4\_1 for the ImageNet dataset, and conv1\_1 and conv2\_1 for the CIFAR-10 and SVHN datasets.
$\mu(\cdot)$ and $\sigma(\cdot)$ denote the mean and the variance, respectively. The Decoder training is to minimize  $\mathcal{L}_{\mathrm{content}}+\mathcal{L}_{\mathrm{style}}$.

\noindent
{\bf Two Feature Space Attacks.}
Recall in the attack phase (Fig.~\ref{fig:attack}), the encoder and decoder are fixed. The style features of a benign image are perturbed while the content features are retained, aiming to trigger misclassification. 
The pre-trained decoder then translates the perturbed embedding 
back to an adversarial sample. 
During perturbation, we focus on minimizing two loss functions.
The first one is the adversarial loss $\mathcal{L}_{M}$ whose goal is to induce misclassification. The second one is similar to the content loss in
the Decoder training (Eq. \ref{eq:content_loss}). Intuitively, although the decoder is trained in a way that it is supposed to decode with minimal loss, arbitrary style perturbation may still cause substantial loss. Hence, such loss has to be considered and minimized during style perturbation.

With two different sets of transformations $\sA$,  we devise two respective kinds of feature space attacks, {\em feature augmentation attack} and {\em feature interpolation attack}. 
For feature augmentation attack, attacker can change both the mean and standard deviation of each channel of the benign embedding independently. The boundary of increments or decrements are set by $\ell_\infty$-norm under logarithmic scale (to achieve stealthiness). Specifically, given two perturbation vectors $\boldsymbol\tau^\mu$ for the mean and $\boldsymbol\tau^\sigma$ for the variance, both have the same dimension $C$ as the embedding (denoting the $C$ channels) and are bounded by $\epsilon$, the list of possible transformations $\sA$ is defined as follows.
\begin{equation}
\begin{split}
\sA  =  \cup_{||\boldsymbol\tau^\sigma||_\infty   \leq \epsilon \mathrm{\;and\;} ||\boldsymbol\tau^\mu||_\infty   \leq \epsilon,\  \boldsymbol\tau^\sigma \mathrm{\;and\;} \boldsymbol\tau^\mu \in \R^{C}}  \\
 \Big \{a :  a(B)_{h,w,c} = e^{\boldsymbol\tau^\sigma_c} (B_{h,w,c} - {\mu_B}_c) + e^{\boldsymbol\tau^\mu_c}{\mu_B}_c \Big \} 
\end{split}
\label{eq:selfaug}
\end{equation}
Note that $\mu_B$ denotes the means of embedding $B$ for the $C$ channels. The subscript $c$ denotes a specific channel. The transformation essentially enlarges the variance of the embedding at channel $c$ by a factor of $e^{\boldsymbol\tau^\sigma_c}$ and the mean by a factor of $e^{\boldsymbol\tau^\mu_c}$.



For the feature interpolation attack, the attacker provides $k$ images 
as the style feature prototypes. Let $\sS_\mu, \sS_\sigma$ be the simplex determined by $\cup_{i \in [1,2,...,k]}{\mu_{f(\vx_i)}}$ and $\cup_{i \in [1,2,...,k]}{\sigma_{f(\vx_i)}}$ respectively. The attacker can modify the vectors of $\mu_B$ and $\sigma_B$ to be any point on the simplex.
\begin{equation}
\begin{split}
\sA & =\cup_{\substack{\sigma_i \in \sS_\sigma \\ \mu_i \in \sS_\mu}}
 \Big \{a : a(B)_{h,w,c} = \sigma_i \cdot \frac{B_{h,w,c} - {\mu_B}_c}{{\sigma_B}_c}  + \mu_i \Big \} 
\end{split}
\label{eq:interpolation}
\end{equation}
Intuitively, it enforces a style constructed from an interpolation of the $k$ style prototypes. Our optimization method is a customized iterative gradient method with gradient clipping (see Appendix~\ref{app:opt}).
In pixel level attacks, two kinds of optimization techniques are widely used: {\em Gradient Sign Method}, e.g., PGD \citep{madry2017towards}, and using continuous function, e.g. tanh, to approximate and bound $\ell_\infty$, e.g., in C\&W \citep{carlini2017towards}.
However in our context, we found these two techniques do not perform well.
Using gradient sign tends to induce a large content loss while using tanh function inside the feature space empirically causes numerical instability.
Instead, we use the iterative gradient method with gradient clipping. Specifically, We first calculate the gradient of loss $\mathcal{L}$ with respect to variables (e.g., $\boldsymbol\tau^\mu_c$ and $\boldsymbol\tau^\sigma_c$ in Equation~(\ref{eq:selfaug})).
The gradient is then clipped by a constant related to the dimension of variables. $||\nabla{\mathcal{L}}||_\infty \leq 10/\sqrt{\mathrm{Dimension\;of\;variable}}$. Then an Adam optimizer iteratively optimizes the variables using the clipped gradients.

For the optimization of Feature Interpolation Attack, style vectors are constrained inside the polygon. To conveniently enforce such constraint, a tensor of variables as coefficients of vertices in the simplex are used to represent the style vectors. These variables are clipped to be positive and the sum of which is normalized to be one during every optimization step. Therefore, these style vectors stay inside the simplex denoted by Equation~(\ref{eq:interpolation}).



\section{Attack Settings}
\label{app:setting}
The two proposed feature space attacks have similar performance on various experimental settings. Unless otherwise stated, we use feature augmentation attack as the default method. For the Encoder, we use VGG-19 from the input layer up to the relu4\_1 for ImageNet, and up to relu2\_1 for CIFAR-10 and SVHN . To launch attacks, we set the $\ell_\infty$-norm of embedding, $\epsilon$, in Eq.~(\ref{eq:selfaug}) to $\ln(1.5)$ for all the untargeted attacks and $\ln(2)$ for all the targeted attacks. We randomly select 1,000 images to perform the attacks on ImageNet.  For CIFAR-10 and SVHN, we use all the inputs in the validation set.

\section{Evaluation}
\label{sec:eval}






Three datasets are employed in the experiments: CIFAR-10 \citep{krizhevsky2009learning}, ImageNet \citep{russakovsky2015imagenet} and SVHN \citep{netzer2011reading}. 
The feature space attack settings can be found in Appendix~\ref{app:setting}.
We use 7 state-of-the-art detection and defense approaches to evaluate the proposed feature space attack. Detection approaches aim to identify adversarial samples while they are provided to a DNN. They often work as an add-on to the model and do not aim to harden the model. 
We use two state-of-the-art adversarial example detection approaches proposed by \cite{roth2019odds} and \cite{dknn} to test our attack. Defense approaches, on the other hand, harden models such that they are robust against adversarial example attacks.
Existing state-of-the-art defense mechanisms either use adversarial training or certify a bound for each input image. We adopt 5 state-of-the-art defense approaches in the literature \citep{madry2017towards,zhang2019theoretically,xie2019feature,song2018improving,lecuyer2019certified} for evaluation. 
Note that while these techniques are intended for pixel space attacks, their effectiveness for our attack is unclear. 
We are unaware of any detection/defense techniques for the kind of feature attacks we are proposing. 
\subsection{Quality of Feature Space Adversarial Examples by Human Study and Distance Metrics}
\label{sec:human_study}

\begin{figure}
    \centering
    \includegraphics[width=.48\textwidth]{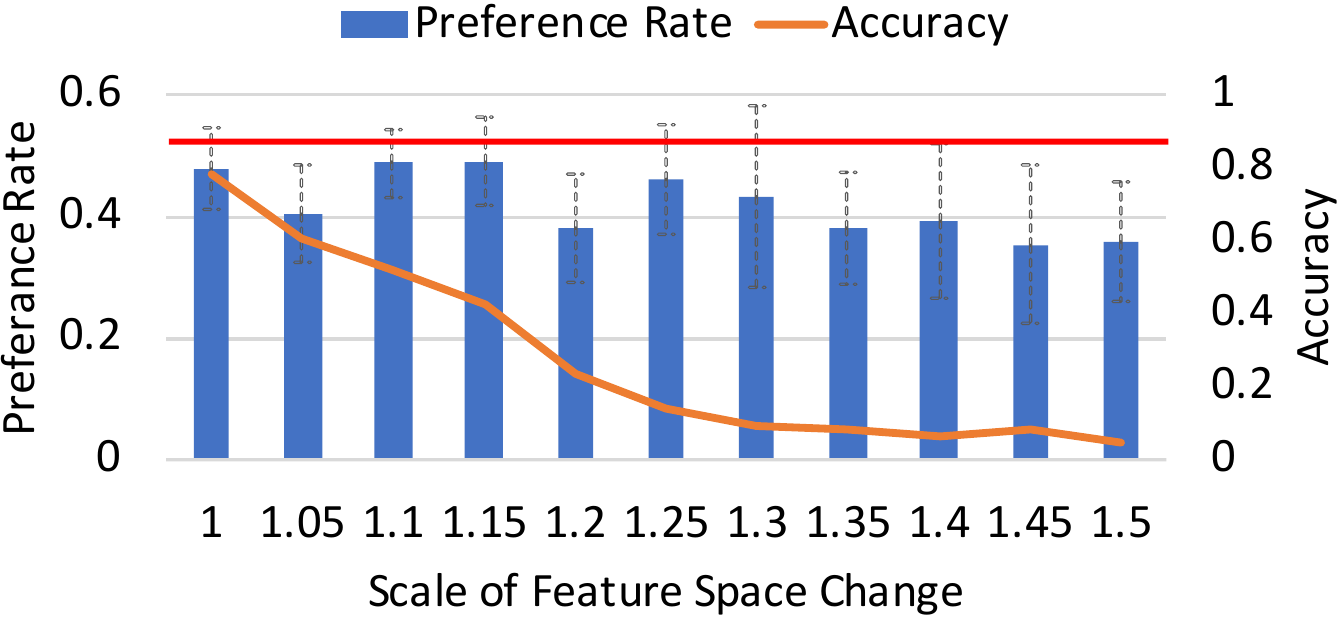}
    \caption{Human preference evaluation. The left y-axis (and blue bar) represents the percentage of user preference towards feature space attack images. The right y-axis (and orange line) denotes the test accuracy of models under feature space attack. 
    The x-axis presents the scale of feature space perturbation $e^\epsilon$ in Eq.~(\ref{eq:selfaug}). The horizontal red line denotes where users cannot distinguish between adversarial samples and original images.}
    \label{fig:human_preference}
\end{figure}

In the first experiment, we conduct a human study to measure the quality of feature space attack samples. We follow the same procedure as in \citep{DBLP:conf/eccv/ZhangIE16,semantic_attack}. Users are given 50 pairs of images, each pair consisting of an original image and its transformed version (by feature space attack). They are asked to choose the realistic one from each pair. The images are randomly selected and used in the following trials. Each pair appears on screen for 3 seconds,
and is evaluated by 10 users. Every user has 5 chances for practice before the trials begin. In total, 110 users completed the study. We repeat the same study for different feature space attack scales on ResNet-50 as shown in Fig.~\ref{fig:human_preference}. On average, 41.9\% of users choose our adversarial samples over original images. This indicates that the feature space attack is largely imperceptible.

We also carry out a set of human studies to qualitatively measure images generated by different attacks: PGD \cite{madry2017towards}, feature space attack, semantic attack \cite{semantic_attack}, HSV attack~\cite{hosseini2018semantic}, manifold attack~\cite{disentangle} and art attack~\cite{artattack}. The results are shown in Table~\ref{tab:compare}. The first columns show the two attacks in comparison. The second column presents the human preference rate. The third column is the attack success rate. In the first table, the two attacks are conducted on a model with the denoise(t,1) defense~\cite{xie2019feature} (in order to avoid 100\% attack success rate). In the following four tables, the attacks are performed on a ResNet-50. We observe that the quality of feature space attack samples is comparable to that of PGD attack and the former has a higher attack success rate.  Feature space attack also outperforms semantic attack, HSV attack, manifold attack and art attack in visual quality while achieving a higher/comparable attack success rate. 
That is, 67\% or more users prefer feature space attack samples to the others. 
The comparison with art attack supports the benefit of leveraging feature-space during attack.
The comparison with manifold attack stresses the importance of manipulating secondary features.
The generated images by these attacks and comparison details can be seen in Appendix \ref{sec:compare_semantic}.

We study the $\ell$-norm distances in both the pixel space and the feature space for both pixel space attacks and feature space attacks. We observe that in the pixel space, the introduced perturbation by feature space attack is much larger than that of the PGD attack. In the feature space, our attack has very similar distances as PGD.
Fig.~\ref{fig:attack_examples} and \ref{fig:images} (in Appendix) show that the adversarial samples have only style differences that are natural or even human imperceptible. Detailed discussion can be found in Appendix~\ref{sec:appendix1}. 
We also study the characteristics of the adversarial samples generated by different feature space attacks and attack settings. Please see Appendix~\ref{sec:appendix2}.

\begin{table}[t]
    \centering
    \scriptsize
    \begin{tabular}{ccc}
    \toprule
              & Pref. & Succ.\\
    \midrule             
    PGD           & 60              & 58              \\
    Feature Space & 40              & 88             \\        
    \bottomrule
    \end{tabular}
    ~~~~~
    \begin{tabular}{ccc}
    \toprule
              & Pref. & Succ.\\
    \midrule             
    Semantic      & 33      & 100              \\
    Feature Space & 67      & 100             \\        
    \bottomrule
    \end{tabular}
    ~~~~~
    \begin{tabular}{ccc}
    \toprule
              & Pref. & Succ.\\
    \midrule             
    HSV      & 20     & 64               \\
    Feature Space & 79  &    100             \\        
    \bottomrule
    \end{tabular}
    ~~~~~
    \begin{tabular}{ccc}
    \toprule
              & Pref. & Succ.\\
    \midrule             
    Manifold      &  27     & 100              \\
    Feature Space &   73    & 100             \\        
    \bottomrule
    \end{tabular}
    ~~~~~
    \begin{tabular}{ccc}
    \toprule
              & Pref. & Succ.\\
    \midrule             
    Art      & 11     & 100               \\
    Feature Space & 89  &    100             \\
    \bottomrule
    \end{tabular}
\caption{Human preference and success rate for different attacks.}
\label{tab:compare}
\end{table}

\subsection{Attack against Detection Approaches}
We use two state-of-the-art adversarial sample detection approaches ``The Odds are Odd'' (O2) \citep{roth2019odds} \footnote{O2 is recently bypassed by \cite{DBLP:journals/corr/abs-1907-12138}, where the attacker already knows the existence of the defense. In our case, however, we are able to evade the detection method without knowing its existence or mechanism.} and feature-space detection method ``Deep k-Nearest Neighbors'' (DkNN) \citep{dknn}.



O2 detects adversarial samples by adding random noise to input images and observing activation changing at a certain layer of a DNN. Specifically, O2 uses the penultimate layer (before the logits layer) as the representation of input images. It then defines a statistical variable that measures pairwise differences between two classes computed from the penultimate layer. The authors observed that adversarial samples differ significantly from benign samples regarding this variable when random noise is added. 
By performing statistical test on this variable, O2 is able to detect PGD attacks \citep{madry2017towards} with over $99\%$ detection rate on CIFAR-10 with bound $\ell_\infty = 8/255$ and on ImageNet with $\ell_\infty = 2/255$. It also has over $90\%$ detection rate against PGD and C\&W \citep{carlini2017towards} attacks under $\ell_2$ metric on CIFAR-10. 



Table~\ref{tab:o2} shows the results of O2 on detecting different input samples. The first two columns are the datasets and models used for evaluation. The third column denotes the prediction accuracy of models on normal inputs. The following three columns present the detection rate of O2 on normal inputs, PGD adversarial samples, and feature space adversarial samples, respectively. The detection rate on normal inputs indicates that O2 falsely recognizes normal inputs as adversarial, which are essentially false positives. We can observe that O2 can effectively detect PGD attack on both datasets, but fails to detect feature space attack. Particularly, O2 has only 0.04\% detection rate on CIFAR-10, which indicates that \textit{O2 can be evaded by feature space attack}. As for ImageNet, O2 can detect 25.30\% of feature space adversarial samples but at the cost of a 19.20\% false positive rate\footnote{The parameters used for ImageNet are not given in the original paper. We can only reduce to this false positive rate after parameter tuning.}. The results show that O2 is ineffective against feature space attack.

Table~\ref{tab:dknn} shows the results for Deep K Nearest Neighbour. Due to memory limits, we only test on the CIFAR-10 dataset. The second column denotes models employed for evaluation including the default one used in the original paper (CNN+MLP). The third column shows model accuracy on benign inputs. The last two columns present detection rate on PGD and feature space attacks. We can observe that DkNN has much lower detection rate on feature space attack compared to PGD, despite the fact that DkNN uses feature space data for detecting adversarial samples.

\begin{table}[t]
    \centering
    \scriptsize
    \begin{tabular}{ c c c c c c}
        \toprule
        \multirow{2}{*}[-.3em]{Dataset} & \multirow{2}{*}[-.3em]{Model} & \multirow{2}{*}[-.3em]{Accuracy} & \multicolumn{3}{c}{Detection Rate} \\
        \cmidrule(lr){4-6}
        &     &     &    Normal  & PGD &  Feature Space  \\
        \midrule
        CIFAR-10 & ResNet-18 & 91.95 & \;\;0.95 & 99.61 & \;\;\textbf{0.04} \\
        ImageNet & ResNet-50 & 75.20 & 19.20 & 99.40 & \textbf{25.30} \\
        \bottomrule
    \end{tabular}
    \caption{O2 detection rate on normal inputs and adversarial samples.}    
    \label{tab:o2}
    \begin{tabular}{ccccc}
        \toprule
        \multirow{2}{*}{Dataset} & \multirow{2}{*}{Model} & \multirow{2}{*}{Accuracy} & \multicolumn{2}{c}{Detection Rate} \\
        \cmidrule(lr){4-5} 
        & & & PGD & Feature Space \\ 
        \midrule
        \multirow{2}{*}{CIFAR-10}
            & CNN+MLP & 53.93 & \;\;3.92 & \textbf{1.95} \\
            & ResNet-18 & 81.51 & 11.32 & \textbf{5.42} \\
        \bottomrule
    \end{tabular}
    \caption{DkNN detection rate on normal inputs and adversarial samples.}    
    \label{tab:dknn}
\end{table}

\begin{table}[t]
    \centering
    \scriptsize
    \begin{tabular}{ccccc}
        \toprule
        \multirow{2}{*}[-.3em]{Attack} & SVHN & \multicolumn{3}{c}{CIFAR-10} \\
        \cmidrule(lr){2-2} \cmidrule(lr){3-5} 
        & Adaption & Madry & TRADES  & Pixel-DP \footnotemark \\
        \midrule
         None               & 84.84 & 77.84 & 84.97 & 44.3  \\
         PGD                & 52.84 & 41.43 & 54.02 & 30.7 \\
         Decoder            & 84.81 & 77.35 & 84.01 & 50.0 \\
         Feature Space      & \;\;\textbf{2.56} & \;\;\textbf{7.05} & \;\;\textbf{8.64} & \;\;\textbf{0.0} \\
        \bottomrule
    \end{tabular}
    \begin{tabular}{cccc}
        \toprule
        \multirow{2}{*}[-.3em]{Attack} & \multicolumn{3}{c}{ImageNet} \\
        \cmidrule(lr){2-4}
        & Denoise (t,1) & Denoise (u,1) & Denoise (u,5) \\
        \midrule
         None               & 61.25 & 61.25 & 78.12 \\
         PGD                & 42.60 & 12.50 & 27.15 \\
         Decoder            & 64.68 & 64.00 & 82.37 \\
         Feature Space      & \textbf{11.41} & \;\;\textbf{1.25} & \;\;\textbf{1.25} \\
        \bottomrule        
         \end{tabular}
    \caption{Evaluation of adversarial attacks against various defense approaches.}    
    \label{tab:attack_eval}
\end{table}


\subsection{Attack against Defense Approaches}
We evaluate our feature space attack on 5 state-of-the-art adversarial training approaches: Madry \citep{madry2017towards}, TRADES \citep{zhang2019theoretically}, Denoise \citep{xie2019feature}, Adaption \citep{song2018improving}, and Pixel-DP \citep{lecuyer2018certified}. For Denoise, the original paper only evaluated on targeted attacks. We conduct experiments on both targeted and untargeted attacks. We use Denoise (t,1) to denote the top-1 accuracy of hardened model on targeted attack and Denoise (u,5) the top-5 accuracy on untargeted attack. We launch the PGD $\ell_\infty$ attack as well as our feature space attack on the four defense approaches. The results are shown in Table~\ref{tab:attack_eval}. 
The first column denotes attack methods, where ``None'' presents the model accuracy on benign inputs and ``Decoder'' denotes the samples directly generated from the decoder without any feature space perturbation. The latter is to show that the Decoder can generate faithful and natural images from embeddings. The following columns show different defense approaches (second row) applied on various datasets (first row). We can see that the PGD attack can reduce model accuracy to some extent when defense mechanisms are in place. \textit{Feature space attack, on the other hand, can effectively reduce model accuracy down to less than 12\%, and most results are one order of magnitude smaller than PGD.} Especially, model accuracy on ImageNet is only 1.25\% when using untargeted attack, even in the presence of the defense technique. 

From the aforementioned results, we observe that existing pixel space detect/defense techniques are largely ineffective as they focus on pixel space. While it may be possible to extend some of these techniques to protect feature space, the needed extension remains unclear to us at this point. We hence leave it to our future work. For example, it is unclear how to extend O2, which leverages the penultimate layer to detect anomaly and hence should have been effective for our attack in theory.

\noindent
{\bf Towards Feature Space Adversarial Training.}
We conduct a preliminary study on using feature space attack to perform adversarial training. 
For comparison, we also perform the PGD adversarial training and use semantic attack to perform adversarial training. 
We evaluate the adversarially trained models against feature space (FS) attack, HSV attack, semantic (SM) attack, and PGD attack, with the first three in the feature space.
We find that PGD adversarial training  is most effective against PGD attack (55\% attack success rate reduction) and has effectiveness against SM attack too (22\% reduction), but not FS or HSV attack. 
Feature space adversarial training can reduce the FS attack success rate by 27\% and the HSV attack by 13\%, but not others. Adversarial training using semantic attack can reduce semantic attack success rate by 34\% and PGD by 38\%, but not others. 
This suggests that different attacks aim at different spaces and the corresponding adversarial trainings may only enhance the corresponding target spaces. 
Note that the robustness improvement of feature space adversarial training is not as substantial as PGD training in the pixel space. We believe that it is because either our study is preliminary and more setups need to be explored; or, feature space adversarial training may be inherently harder and demand new methods. We will leave it to our future study. More details (e.g., ablation study) can be found in Appendix~\ref{app:advtraining}.

\section{Conclusion}

We propose feature space adversarial attack on DNNs. It is based on perturbing style features and retaining content features. Such attacks inject natural style changes to input images to cause model misclassification. Since they usually cause substantial pixel space perturbations and existing detection/defense techniques are mostly for bounded pixel space attacks, these techniques are not effective for feature space attacks.


\section{Acknowledgments}
This research was supported, in part by NSF 1901242 and 1910300, ONR N000141712045, N000141410468 and N000141712947, and IARPA TrojAI W911NF-19-S-0012. Any opinions, findings, and conclusions in this paper are those of the authors only and do not necessarily reflect the views of our sponsors.

\onecolumn

\bibliography{reference}

\begin{thebibliography}{34}
\providecommand{\natexlab}[1]{#1}
\providecommand{\url}[1]{\texttt{#1}}
\expandafter\ifx\csname urlstyle\endcsname\relax
  \providecommand{\doi}[1]{doi: #1}\else
  \providecommand{\doi}{doi: \begingroup \urlstyle{rm}\Url}\fi

\bibitem[Szegedy et~al.(2014)Szegedy, Zaremba, Sutskever, Bruna, Erhan,
  Goodfellow, and Fergus]{szegedy2013intriguing}
Christian Szegedy, Wojciech Zaremba, Ilya Sutskever, Joan Bruna, Dumitru Erhan,
  Ian Goodfellow, and Rob Fergus.
\newblock {Intriguing Properties of Neural Networks}.
\newblock In \emph{International Conference on Learning Representations
  (ICLR)}, 2014.

\bibitem[Carlini and Wagner(2017)]{carlini2017towards}
Nicholas Carlini and David Wagner.
\newblock Towards evaluating the robustness of neural networks.
\newblock In \emph{Proceedings of 38th IEEE Symposium on Security and Privacy
  (SP)}, pages 39--57. IEEE, 2017.

\bibitem[Qin et~al.(2019)Qin, Carlini, Goodfellow, Cottrell, and
  Raffel]{qin2019imperceptible}
Yao Qin, Nicholas Carlini, Ian Goodfellow, Garrison Cottrell, and Colin Raffel.
\newblock {Imperceptible, Robust, and Targeted Adversarial Examples for
  Automatic Speech Recognition}.
\newblock In \emph{Proceedings of the 36th International Conference on Machine
  Learning (ICML)}, pages 5231--5240, 2019.

\bibitem[Ebrahimi et~al.(2018)Ebrahimi, Rao, Lowd, and
  Dou]{ebrahimi2018hotflip}
Javid Ebrahimi, Anyi Rao, Daniel Lowd, and Dejing Dou.
\newblock Hotflip: White-box adversarial examples for text classification.
\newblock In \emph{Proceedings of the 56th Annual Meeting of the Association
  for Computational Linguistics (ACL)}, pages 31--36, 2018.

\bibitem[Li et~al.(2019)Li, Neupane, Paul, Song, Krishnamurthy, Chowdhury, and
  Swami]{li2018adversarial}
Shasha Li, Ajaya Neupane, Sujoy Paul, Chengyu Song, Srikanth~V Krishnamurthy,
  Amit K~Roy Chowdhury, and Ananthram Swami.
\newblock Adversarial perturbations against real-time video classification
  systems.
\newblock In \emph{Proceedings of 26th Annual Network and Distributed System
  Security Symposium (NDSS)}, 2019.

\bibitem[Tao et~al.(2018)Tao, Ma, Liu, and Zhang]{tao2018attacks}
Guanhong Tao, Shiqing Ma, Yingqi Liu, and Xiangyu Zhang.
\newblock Attacks meet interpretability: Attribute-steered detection of
  adversarial samples.
\newblock In \emph{Advances in Neural Information Processing Systems
  (NeurIPS)}, pages 7717--7728, 2018.

\bibitem[Ma et~al.(2019)Ma, Liu, Tao, Lee, and Zhang]{ma2019nic}
Shiqing Ma, Yingqi Liu, Guanhong Tao, Wen-Chuan Lee, and Xiangyu Zhang.
\newblock Nic: Detecting adversarial samples with neural network invariant
  checking.
\newblock In \emph{Proceedings of 26th Annual Network and Distributed System
  Security Symposium (NDSS)}, 2019.

\bibitem[Madry et~al.(2018)Madry, Makelov, Schmidt, Tsipras, and
  Vladu]{madry2017towards}
Aleksander Madry, Aleksandar Makelov, Ludwig Schmidt, Dimitris Tsipras, and
  Adrian Vladu.
\newblock Towards deep learning models resistant to adversarial attacks.
\newblock In \emph{Proceedings of 6th International Conference on Learning
  Representations (ICLR)}, 2018.

\bibitem[Zhang et~al.(2019)Zhang, Yu, Jiao, Xing, El~Ghaoui, and
  Jordan]{zhang2019theoretically}
Hongyang Zhang, Yaodong Yu, Jiantao Jiao, Eric Xing, Laurent El~Ghaoui, and
  Michael Jordan.
\newblock Theoretically principled trade-off between robustness and accuracy.
\newblock In \emph{International Conference on Machine Learning (ICML)}, pages
  7472--7482, 2019.

\bibitem[Huang and Belongie(2017)]{huang2017arbitrary}
Xun Huang and Serge Belongie.
\newblock Arbitrary style transfer in real-time with adaptive instance
  normalization.
\newblock In \emph{Proceedings of the IEEE International Conference on Computer
  Vision (ICCV)}, pages 1501--1510, 2017.

\bibitem[Hosseini and Poovendran(2018)]{hosseini2018semantic}
Hossein Hosseini and Radha Poovendran.
\newblock Semantic adversarial examples.
\newblock In \emph{Proceedings of the IEEE Conference on Computer Vision and
  Pattern Recognition Workshops}, pages 1614--1619, 2018.

\bibitem[Bhattad et~al.(2020)Bhattad, Chong, Liang, Li, and
  Forsyth]{semantic_attack}
Anand Bhattad, Minjin Chong, Kaizhao Liang, Bo~Li, and David Forsyth.
\newblock Unrestricted adversarial examples via semantic manipulation.
\newblock In \emph{International Conference on Learning Representations {ICLR}
  Conference, 2020}, 2020.

\bibitem[Roth et~al.(2019)Roth, Kilcher, and Hofmann]{roth2019odds}
Kevin Roth, Yannic Kilcher, and Thomas Hofmann.
\newblock The odds are odd: A statistical test for detecting adversarial
  examples.
\newblock In \emph{International Conference on Machine Learning (ICML)}, pages
  5498--5507, 2019.

\bibitem[Xie et~al.(2019)Xie, Wu, Maaten, Yuille, and He]{xie2019feature}
Cihang Xie, Yuxin Wu, Laurens van~der Maaten, Alan~L Yuille, and Kaiming He.
\newblock Feature denoising for improving adversarial robustness.
\newblock In \emph{Proceedings of the IEEE Conference on Computer Vision and
  Pattern Recognition (CVPR)}, pages 501--509, 2019.

\bibitem[Gatys et~al.(2016)Gatys, Ecker, and Bethge]{gatys2016image}
Leon~A Gatys, Alexander~S Ecker, and Matthias Bethge.
\newblock Image style transfer using convolutional neural networks.
\newblock In \emph{Proceedings of the IEEE Conference on Computer Vision and
  Pattern Recognition (CVPR)}, pages 2414--2423, 2016.

\bibitem[Li and Wand(2016)]{li2016precomputed}
Chuan Li and Michael Wand.
\newblock Precomputed real-time texture synthesis with markovian generative
  adversarial networks.
\newblock In \emph{European Conference on Computer Vision (ECCV)}, pages
  702--716, 2016.

\bibitem[Johnson et~al.(2016)Johnson, Alahi, and
  Fei-Fei]{johnson2016perceptual}
Justin Johnson, Alexandre Alahi, and Li~Fei-Fei.
\newblock Perceptual losses for real-time style transfer and super-resolution.
\newblock In \emph{European Conference on Computer Vision (ECCV)}, pages
  694--711, 2016.

\bibitem[Dumoulin et~al.(2017)Dumoulin, Shlens, and
  Kudlur]{dumoulin2016learned}
Vincent Dumoulin, Jonathon Shlens, and Manjunath Kudlur.
\newblock A learned representation for artistic style.
\newblock In \emph{Proceedings of 5th International Conference on Learning
  Representations (ICLR)}, 2017.

\bibitem[Li et~al.(2017)Li, Fang, Yang, Wang, Lu, and Yang]{li2017diversified}
Yijun Li, Chen Fang, Jimei Yang, Zhaowen Wang, Xin Lu, and Ming-Hsuan Yang.
\newblock Diversified texture synthesis with feed-forward networks.
\newblock In \emph{Proceedings of the IEEE Conference on Computer Vision and
  Pattern Recognition (CVPR)}, pages 3920--3928, 2017.

\bibitem[Inkawhich et~al.(2019)Inkawhich, Wen, Li, and Chen]{transfer}
Nathan Inkawhich, Wei Wen, Hai Li, and Yiran Chen.
\newblock Feature space perturbations yield more transferable adversarial
  examples.
\newblock pages 7059--7067, 06 2019.
\newblock \doi{10.1109/CVPR.2019.00723}.

\bibitem[Laidlaw and Feizi(2019)]{laidlaw2019functional}
Cassidy Laidlaw and Soheil Feizi.
\newblock Functional adversarial attacks.
\newblock In \emph{Advances in Neural Information Processing Systems}, pages
  10408--10418, 2019.

\bibitem[Prabhu and UnifyID(2018)]{artattack}
Vinay~Uday Prabhu and John~Whaley UnifyID.
\newblock Art-attack ! on style transfers with textures , label categories and
  adversarial examples.
\newblock 2018.

\bibitem[Song et~al.(2018)Song, Shu, Kushman, and Ermon]{song2018constructing}
Yang Song, Rui Shu, Nate Kushman, and Stefano Ermon.
\newblock Constructing unrestricted adversarial examples with generative
  models.
\newblock In \emph{Advances in Neural Information Processing Systems}, pages
  8312--8323, 2018.

\bibitem[Stutz et~al.(2018)Stutz, Hein, and Schiele]{disentangle}
David Stutz, Matthias Hein, and Bernt Schiele.
\newblock Disentangling adversarial robustness and generalization.
\newblock \emph{CoRR}, abs/1812.00740, 2018.
\newblock URL \url{http://arxiv.org/abs/1812.00740}.

\bibitem[Krizhevsky et~al.(2009)]{krizhevsky2009learning}
Alex Krizhevsky et~al.
\newblock Learning multiple layers of features from tiny images.
\newblock Technical report, Citeseer, 2009.

\bibitem[Russakovsky et~al.(2015)Russakovsky, Deng, Su, Krause, Satheesh, Ma,
  Huang, Karpathy, Khosla, Bernstein, et~al.]{russakovsky2015imagenet}
Olga Russakovsky, Jia Deng, Hao Su, Jonathan Krause, Sanjeev Satheesh, Sean Ma,
  Zhiheng Huang, Andrej Karpathy, Aditya Khosla, Michael Bernstein, et~al.
\newblock Imagenet large scale visual recognition challenge.
\newblock \emph{International Journal of Computer Vision}, 115\penalty0
  (3):\penalty0 211--252, 2015.

\bibitem[Netzer et~al.(2011)Netzer, Wang, Coates, Bissacco, Wu, and
  Ng]{netzer2011reading}
Yuval Netzer, Tao Wang, Adam Coates, Alessandro Bissacco, Bo~Wu, and Andrew~Y
  Ng.
\newblock Reading digits in natural images with unsupervised feature learning.
\newblock In \emph{NIPS Workshop on Deep Learning and Unsupervised Feature
  Learning}, 2011.

\bibitem[Papernot and McDaniel(2018)]{dknn}
Nicolas Papernot and Patrick~D. McDaniel.
\newblock Deep k-nearest neighbors: Towards confident, interpretable and robust
  deep learning.
\newblock \emph{CoRR}, abs/1803.04765, 2018.
\newblock URL \url{http://arxiv.org/abs/1803.04765}.

\bibitem[Song et~al.(2019)Song, He, Wang, and Hopcroft]{song2018improving}
Chuanbiao Song, Kun He, Liwei Wang, and John~E. Hopcroft.
\newblock Improving the generalization of adversarial training with domain
  adaptation.
\newblock In \emph{International Conference on Learning Representations
  (ICLR)}, 2019.

\bibitem[Lecuyer et~al.(2019{\natexlab{a}})Lecuyer, Atlidakis, Geambasu, Hsu,
  and Jana]{lecuyer2019certified}
Mathias Lecuyer, Vaggelis Atlidakis, Roxana Geambasu, Daniel Hsu, and Suman
  Jana.
\newblock Certified robustness to adversarial examples with differential
  privacy.
\newblock In \emph{2019 IEEE Symposium on Security and Privacy (SP)}, pages
  656--672. IEEE, 2019{\natexlab{a}}.

\bibitem[Zhang et~al.(2016)Zhang, Isola, and Efros]{DBLP:conf/eccv/ZhangIE16}
Richard Zhang, Phillip Isola, and Alexei~A. Efros.
\newblock Colorful image colorization.
\newblock In Bastian Leibe, Jiri Matas, Nicu Sebe, and Max Welling, editors,
  \emph{Computer Vision - {ECCV} 2016 - 14th European Conference, Amsterdam,
  The Netherlands, October 11-14, 2016, Proceedings, Part {III}}, volume 9907
  of \emph{Lecture Notes in Computer Science}, pages 649--666. Springer, 2016.
\newblock \doi{10.1007/978-3-319-46487-9\_40}.
\newblock URL \url{https://doi.org/10.1007/978-3-319-46487-9\_40}.

\bibitem[Hosseini et~al.(2019)Hosseini, Kannan, and
  Poovendran]{DBLP:journals/corr/abs-1907-12138}
Hossein Hosseini, Sreeram Kannan, and Radha Poovendran.
\newblock Are odds really odd? bypassing statistical detection of adversarial
  examples.
\newblock \emph{CoRR}, abs/1907.12138, 2019.
\newblock URL \url{http://arxiv.org/abs/1907.12138}.

\bibitem[Lecuyer et~al.(2019{\natexlab{b}})Lecuyer, Atlidakis, Geambasu, Hsu,
  and Jana]{lecuyer2018certified}
Mathias Lecuyer, Vaggelis Atlidakis, Roxana Geambasu, Daniel Hsu, and Suman
  Jana.
\newblock Certified robustness to adversarial examples with differential
  privacy.
\newblock In \emph{Proceedings of 40th IEEE Symposium on Security and Privacy
  (SP)}, 2019{\natexlab{b}}.

\bibitem[Zagoruyko and Komodakis(2016)]{DBLP:journals/corr/ZagoruykoK16}
Sergey Zagoruyko and Nikos Komodakis.
\newblock Wide residual networks.
\newblock \emph{CoRR}, abs/1605.07146, 2016.
\newblock URL \url{http://arxiv.org/abs/1605.07146}.

\end{thebibliography}

\newpage
\appendix
\onecolumn
\title{Appendix}


%
\def\toptitlebar{
	\hrule height4pt
	}

\def\bottomtitlebar{
	\vskip .2in
	\hrule height1pt
	\vskip .1in
	}

\thispagestyle{empty}
\hsize\textwidth
\linewidth\hsize \toptitlebar\vspace{+0.1in} \begin{center}
    {
{\large\bf Appendix \par}}
\end{center} 
\vspace{-0.1in} \bottomtitlebar

\section{Optimization}
\label{app:opt}

\section{Measurement of Perturbation in Pixel and Feature Spaces}
\label{sec:appendix1}
To measure the magnitude of perturbation introduced by adversarial attacks, we use both $\ell_\infty$ and $\ell_2$ distances. In addition, as we aim to understand how the different attacks perturb the pixel space and the feature space, we compute the distances for both spaces. For the pixel space, the calculation is discussed in Backgrond section. For the feature space, 
we normalize the embeddings before distance calculation. For each channel, we use $h(\vx)= \frac{f(\vx)-\mu_{f(\vx)}}{\sigma_{f(\vx)}}$ to normalize the embedding produced by the Encoder $f(\cdot)$ given input $\vx$. The feature space difference hence can be computed as $||h(\vx)-h(\vx')||_p$. Table~\ref{tab:perturb_svhn}, Table~\ref{tab:perturb_cifar} and Table~\ref{tab:perturb_imagenet} illustrate the magnitude of perturbation introduced by adversarial attacks on pixel and feature spaces for different hardened models. It can be observed that in the pixel space, the introduced perturbation by feature space attack is much larger than that of the PGD attack with $\ell_\infty$ and $\ell_2$ distances. In the feature space, however, our feature space attack does not induce large difference between normal inputs and adversarial samples. Particularly, the difference is similar or even smaller than that by the PGD attack. 
As shown in Fig.~\ref{fig:attack_examples} and Fig.~\ref{fig:images}, the introduced perturbation is either insensitive to humans or even imperceptible.

\begin{table}[h]
    \centering
    \begin{tabular}{ccccc}
        \toprule
        \multirow{2}{*}[-.3em]{Attack} & \multicolumn{2}{c}{Pixel Space} & \multicolumn{2}{c}{Feature Space} \\
        \cmidrule(lr){2-3} \cmidrule(lr){4-5}
        & $l_\infty$ & $l_2$ & $l_\infty$ & $l_2$ \\
        \midrule
        PGD             & 0.02 & 1.04 & 11.50 & 53.78 \\
        Decoder         & 0.08 & 1.05 & 10.12 & 36.52 \\
        Feature Space   & 0.12 & 2.01 & 10.51 & 41.83 \\
        \bottomrule
    \end{tabular}
    \caption{Magnitude of perturbation on hardened SVHN models.}
    \label{tab:perturb_svhn}
\end{table}

\begin{table}[h]
    \centering
    \begin{tabular}{ccccccccc}
        \toprule
        \multirow{3}{*}[-.5em]{Attack} & \multicolumn{4}{c}{Madry} & \multicolumn{4}{c}{TRADES} \\
        \cmidrule(lr){2-5} \cmidrule(lr){6-9}
        & \multicolumn{2}{c}{Pixle Space} & \multicolumn{2}{c}{Feature Space} & \multicolumn{2}{c}{Pixle Space} & \multicolumn{2}{c}{Feature Space} \\
        \cmidrule(lr){2-3} \cmidrule(lr){4-5} \cmidrule(lr){6-7} \cmidrule(lr){8-9}
        & $\ell_\infty$ & $\ell_2$ & $\ell_\infty$ & $\ell_2$ & $\ell_\infty$ & $\ell_2$ & $\ell_\infty$ & $\ell_2$ \\
        \midrule
         PGD            & 0.03 & 1.51 & 6.86 & 41.71
                        & 0.03 & 1.47 & 6.49 & 37.89 \\
         Decoder        & 0.19 & 1.85 & 4.29 & 25.13
                        & 0.18 & 1.85 & 4.30 & 25.15 \\
         Feature Space  & 0.27 & 4.08 & 6.88 & 43.38
                        & 0.28 & 4.72 & 7.43 & 46.43 \\
        \bottomrule
        \end{tabular}
    \caption{Magnitude of perturbation on hardened CIFAR-10 models.}
    \label{tab:perturb_cifar}
\end{table}

\begin{table}[h]
    \centering
    \begin{tabular}{ccccccccc}
        \toprule
        \multirow{3}{*}[-.5em]{Attack} & \multicolumn{4}{c}{Targeted} & \multicolumn{4}{c}{Untargeted} \\
        \cmidrule(lr){2-5} \cmidrule(lr){6-9}
        & \multicolumn{2}{c}{Pixle Space} & \multicolumn{2}{c}{Feature Space} & \multicolumn{2}{c}{Pixle Space} & \multicolumn{2}{c}{Feature Space} \\
        \cmidrule(lr){2-3} \cmidrule(lr){4-5} \cmidrule(lr){6-7} \cmidrule(lr){8-9}
        & $\ell_\infty$ & $\ell_2$ & $\ell_\infty$ & $\ell_2$ & $\ell_\infty$ & $\ell_2$ & $\ell_\infty$ & $\ell_2$ \\
        \midrule
         PGD            & 0.06 & 19.48 & 12.08 & 375
                        & 0.03 & \;\;9.24 & 16.91 & 227 \\
         Decoder        & 0.87 & 44.98 & \;\;8.26 & 193
                        & 0.88 & 44.67 & 15.24 & 214 \\
         Feature Space  & 0.89 & 69.07 & \;\;9.99 & 283
                        & 0.86 & 54.89 & 16.09 & 236 \\
        \bottomrule
        \end{tabular}
    \caption{Magnitude of perturbation on hardened ImageNet models.}    
    \label{tab:perturb_imagenet}
\end{table}

\section{Two Feature Space Attacks}
\label{sec:appendix2}
\begin{figure}[h]
    \centering
    \includegraphics[width=1\textwidth]{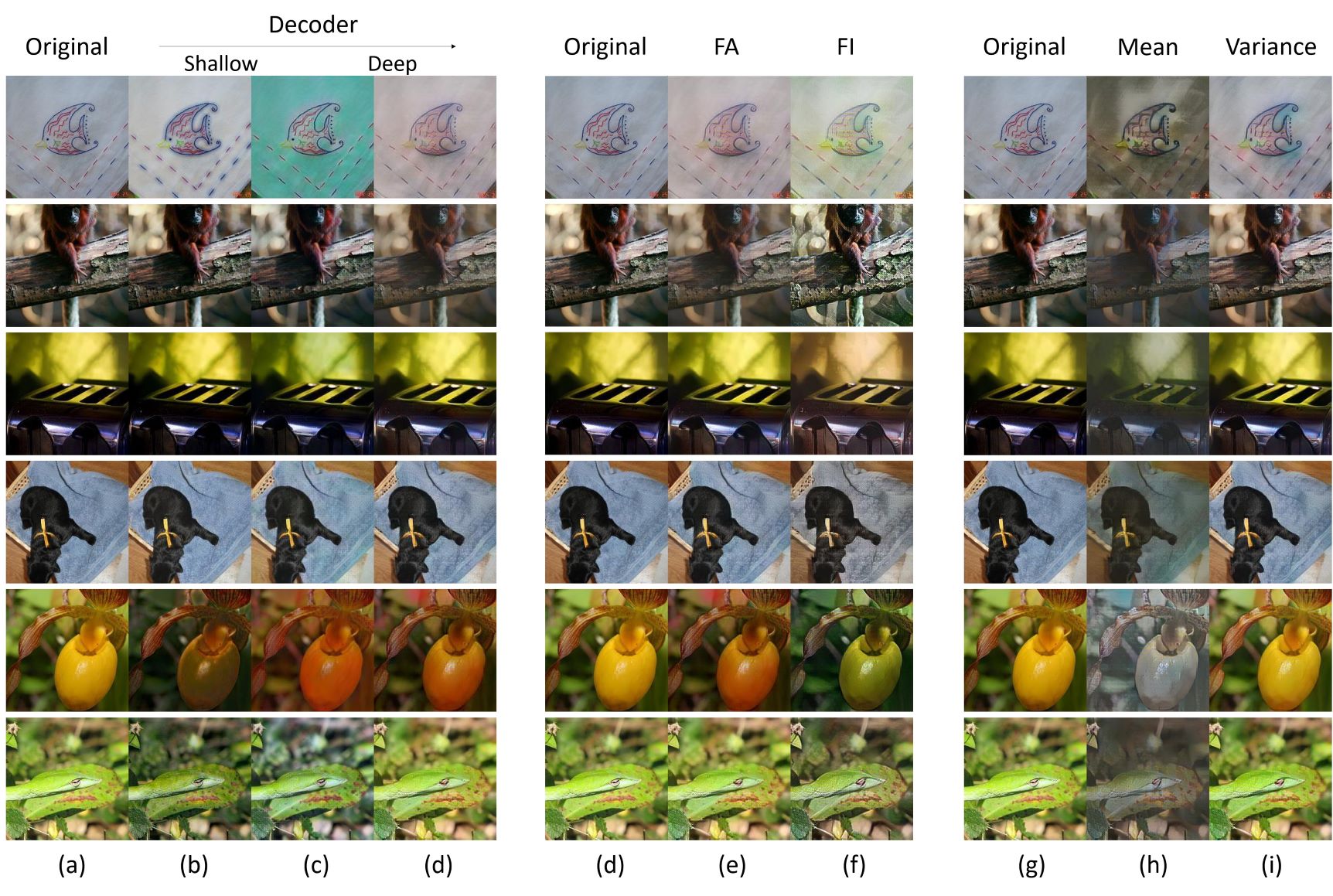}
    \caption{The adversarial samples from different feature space attack methods.}
    \label{fig:images}
\end{figure}

We generate and visually analyze the two feature space attacks. The ImageNet model hardened by feature denoising is used for generating adversarial samples. Columns (a), (d), and (g) in Fig.~\ref{fig:images} present the original images. Columns (b), (c), and (d) present the adversarial samples generated by the Encoder and the corresponding Decoder, with different encoder depths. Specifically, column (b) uses conv2\_1 layers, column (c) uses conv3\_1 layers and column (d) uses conv4\_1 layers of a pre-trained VGG-19 as the Encoder, and the Decoders are of neural network structure similar to the corresponding Encoders but in a reverse order. Observe that as the Encoder becomes  deeper, object outlines and textures are changed in addition to colors. 
Columns (e) and (f) are adversarial samples generated by the feature argumentation attack (FA) and feature interpolation attack (FI). We observe that they both generate realistic images.

In column (h), we only perturb the mean of embedding whereas in column (i) we only perturb the standard deviation of embedding. The results indicate that mean values tend to represent the background and the overall color tone. In contrast, the standard deviations tend to represent the object shape and relative color. 

\section{Comparison with Other Attacks}
\label{sec:compare_semantic}
\begin{figure}[ht!]
    \centering
    \begin{subfigure}[t]{.115\textwidth}
        \centering
        \includegraphics[width=\textwidth]{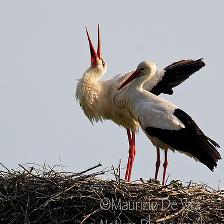}
        \includegraphics[width=\textwidth]{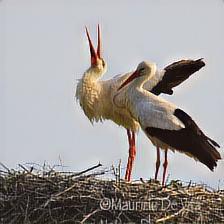}
        \includegraphics[width=\textwidth]{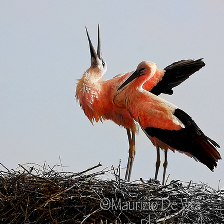}
        \includegraphics[width=\textwidth]{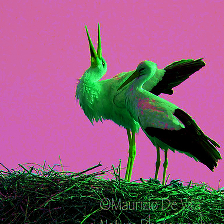}
        \includegraphics[width=\textwidth]{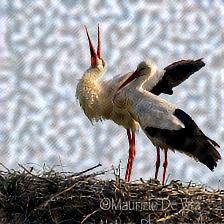}
        \includegraphics[width=\textwidth]{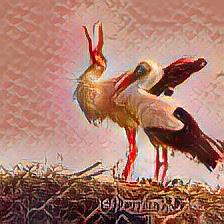}
        \captionsetup{justification=centering}
        \caption{}
        \label{fig:my_label}
    \end{subfigure}    
    \begin{subfigure}[t]{.115\textwidth}
        \centering
        \includegraphics[width=\textwidth]{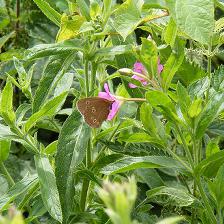}
        \includegraphics[width=\textwidth]{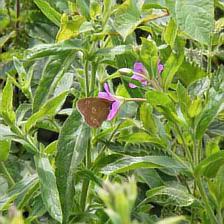}
        \includegraphics[width=\textwidth]{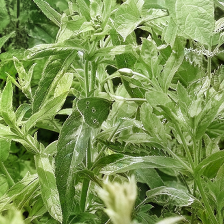}
        \includegraphics[width=\textwidth]{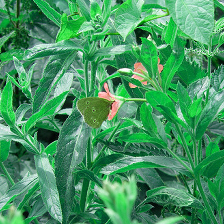}
        \includegraphics[width=\textwidth]{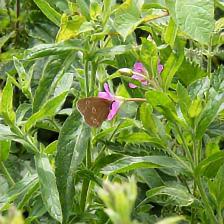}
        \includegraphics[width=\textwidth]{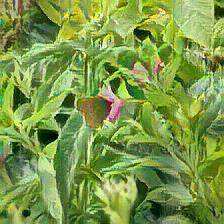}
        \caption{}
        \label{fig:my_label}
    \end{subfigure}
    \begin{subfigure}[t]{.115\textwidth}
        \centering
        \includegraphics[width=\textwidth]{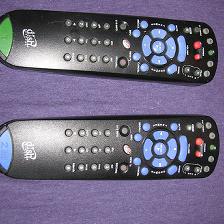}
        \includegraphics[width=\textwidth]{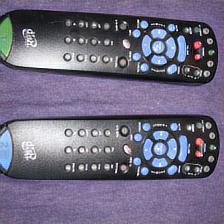}
        \includegraphics[width=\textwidth]{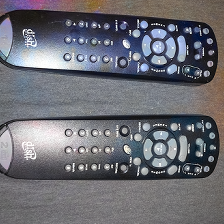}
        \includegraphics[width=\textwidth]{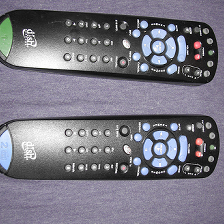}
        \includegraphics[width=\textwidth]{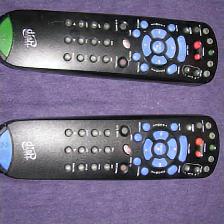}
        \includegraphics[width=\textwidth]{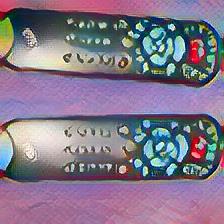}
        \captionsetup{justification=centering}
        \caption{}
        \label{fig:my_label}
    \end{subfigure}
    \begin{subfigure}[t]{.115\textwidth}
        \centering
        \includegraphics[width=\textwidth]{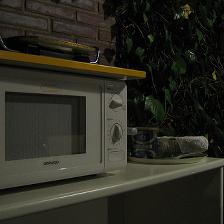}
        \includegraphics[width=\textwidth]{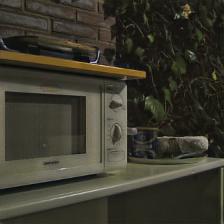}
        \includegraphics[width=\textwidth]{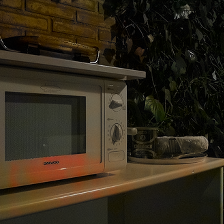}
        \includegraphics[width=\textwidth]{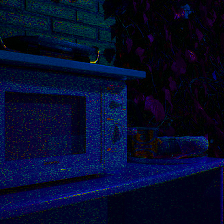}
        \includegraphics[width=\textwidth]{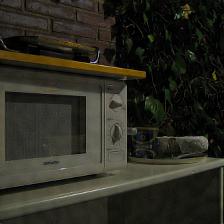}
        \includegraphics[width=\textwidth]{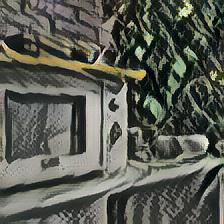}
        \captionsetup{justification=centering}
        \caption{}
        \label{fig:my_label}
    \end{subfigure}
    \begin{subfigure}[t]{.115\textwidth}
        \centering
        \includegraphics[width=\textwidth]{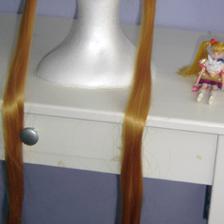}
        \includegraphics[width=\textwidth]{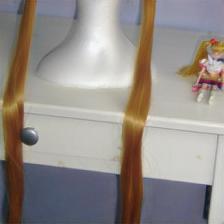}
        \includegraphics[width=\textwidth]{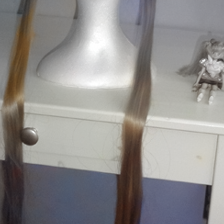}
        \includegraphics[width=\textwidth]{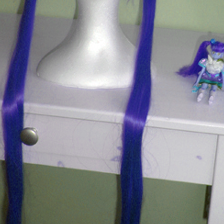}
        \includegraphics[width=\textwidth]{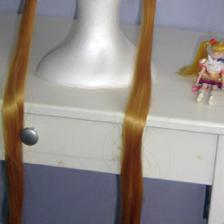}
        \includegraphics[width=\textwidth]{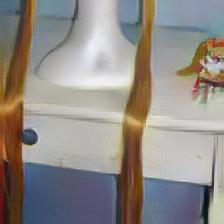}
        \captionsetup{justification=centering}
        \caption{}
        \label{fig:my_label}
    \end{subfigure}
    \begin{subfigure}[t]{.115\textwidth}
        \centering
        \includegraphics[width=\textwidth]{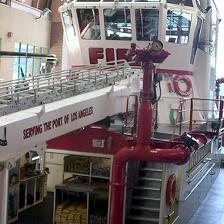}
        \includegraphics[width=\textwidth]{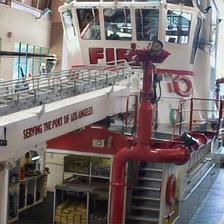}
        \includegraphics[width=\textwidth]{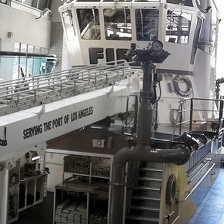}
        \includegraphics[width=\textwidth]{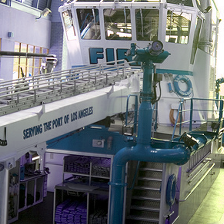}
        \includegraphics[width=\textwidth]{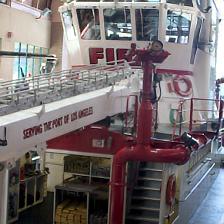}
        \includegraphics[width=\textwidth]{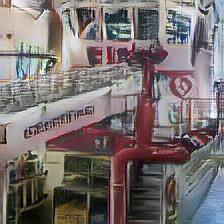}
        \captionsetup{justification=centering}
        \caption{}
        \label{fig:my_label}
    \end{subfigure}
     \begin{subfigure}[t]{.115\textwidth}
        \centering
        \includegraphics[width=\textwidth]{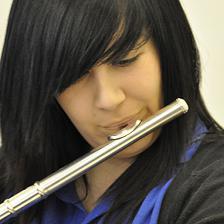}
        \includegraphics[width=\textwidth]{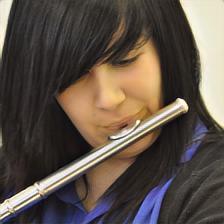}
        \includegraphics[width=\textwidth]{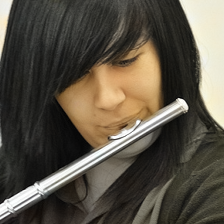}
        \includegraphics[width=\textwidth]{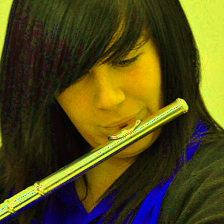}
        \includegraphics[width=\textwidth]{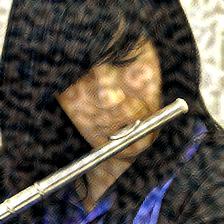}
        \includegraphics[width=\textwidth]{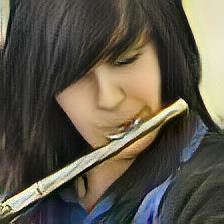}
        \captionsetup{justification=centering}
        \caption{}
        \label{fig:my_label}
    \end{subfigure}
    \begin{subfigure}[t]{.115\textwidth}
        \centering
        \includegraphics[width=\textwidth]{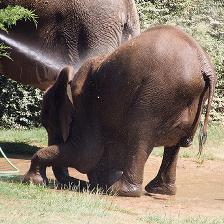}
        \includegraphics[width=\textwidth]{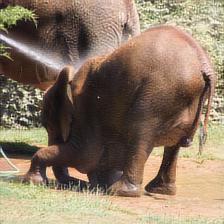}
        \includegraphics[width=\textwidth]{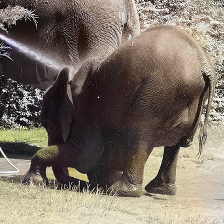}
        \includegraphics[width=\textwidth]{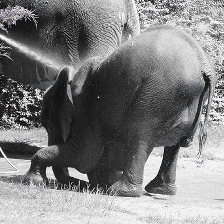}
        \includegraphics[width=\textwidth]{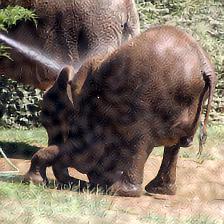}
        \includegraphics[width=\textwidth]{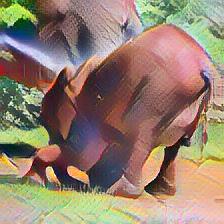}
        \captionsetup{justification=centering}
        \caption{}
        \label{fig:my_label}
    \end{subfigure}
    \caption{Examples by feature space attack, semantic attack, HSV attack, manifold attack and art attack. The first row presents the original images. The second row shows the adversarial images generated by feature space attack. The third row denotes the adversarial samples produced by semantic attack. The fourth row shows the adversarial examples by HSV attack. The fifth row shows the results by manifold attack. And the last row shows the adversarial samples by art attack.}
    \label{fig:attack_comparison}
\end{figure}

Some examples used in the human studies for comparing between feature space attack and semantic attack, HSV attack, manifold attack and art attack are shown in Fig.~\ref{fig:attack_comparison}. The top row presents the original images. The second row shows the adversarial images generated by feature space attack. The third row denotes the adversarial samples produced by the semantic attack. And the last row shows the adversarial examples by the HSV attack. We can observe that the adversarial samples generated by the feature space attack look more natural.


In the following, we introduce the settings used for comparing with other attacks. We adopt the source code of these attacks and utilize the best settings stated in their papers. Specifically, for the HSV attack, we randomly sample 1000 times and select the first successful adversarial samples. For the semantic attack, we use their provided color model and the setting cadv$_8$ in the original paper. Since the semantic attack is unbounded, we use the first successful adversarial sample during the optimization procedure, which is the most similar one to the original image. 
For the art attack,  we gradually increase the interpolation coefficient as defined in the original paper and select the first successful adversarial sample. We repeat this process on the multiple art works provided by the authors. 
The manifold attack paper does not include a suitable setting to scale up to ImageNet. Thus we compare with it using the same network structure as ours. We set the internal $\ell_\infty$ bound in a similar manner as in the paper.

\section{Towards Feature Space Adversarial Training}
\label{app:advtraining}

\begin{table}[h]
\centering
\begin{tabular}{cccccc}
\toprule
\multirow{2}{*}[-0.3em]{Model} & \multirow{2}{*}[-0.3em]{Accuracy} & \multicolumn{4}{c}{Success Rates}                           \\ \cmidrule(lr){3-6} 
                       &                           & FS  & HSV & SM & PGD   \\ 
\midrule
Normal                 & \textbf{92.1}\%                   & 100\%   & 62.2\%    & 100\%                      & 99.6\% \\ 
PGD-Adv        & 78.1\%                   & 94.0\% & 65.1\%    & 78.8\%                    & \textbf{44.2}\% \\ 
SM-Adv          & 76.9\%                   & 95.7\% & 69.9\%    & \textbf{66.2}\%                    & 62.5\% \\ 
FS-Adv         & 82.4\%                   & \textbf{73.3}\% & \textbf{48.3}\%    & 92.7\%                    & 92.7\% \\
\bottomrule
\end{tabular}
\caption{ Adversarial training results
}
\label{tab:adv_train}
\end{table}


       


\begin{table}[h]
\centering
\begin{tabular}{cccc}
\toprule
Model        & Steps & Normal Accuracy & Feature Space Attack Success Rates \\
\midrule
ResNet18     & 50    & 82.1\% & 83.2\%                             \\
ResNet18     & 100   & 82.4\% & 78.0\%                             \\
ResNet18     & 200   & 81.9\% & 73.7\%                             \\
ResNet18     & 400   & 82.4\% & 73.3\%                             \\
WResNet18 & 200   & 82.9\% &    72.5\%                                  \\
ResNet28     & 200   &  84.5\%  &  66.4\%       \\
\bottomrule
\end{tabular}
\caption{Ablation study for feature space adversarial training}
\label{tab:adv_ablation}
\end{table}

\begin{figure}[h]
    \centering
    \includegraphics[width=.6\textwidth]{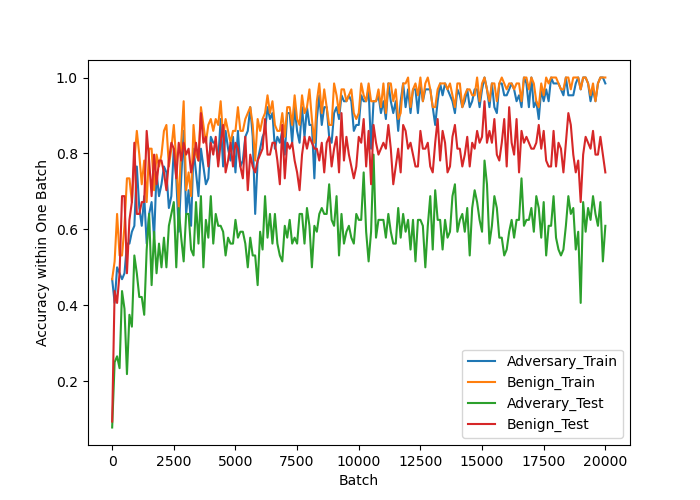}
    \caption{Learning curve of adversarial training on ResNet18, with the number steps of feature space attack set to 400. We show the accuracy in the presence of feature space adversary and the normal accuracy of the model throughout the training process, on both the training set and the test set.
    }
    \label{fig:learningcurve}
\end{figure}

The ultimate goal of developing new adversarial attack is to enhance model robustness. In this section, we present a preliminary study on using feature space attack to adversarial-train the model. 
For comparison, we also perform the PGD adversarial training and  use semantic attack to perform adversarial training. 
We evaluate the adversarially trained models against feature space (FS) attack, HSV attack, semantic (SM) attack, and PGD attack, with the first three in the feature space.

We adopt a standard adversarial training procedure as in~\cite{carlini2017towards}, where adversarial samples are generated against the model and used to train the classifier iteratively. We conduct the experiment on the CIFAR-10 dataset and the ResNet-18 model. For semantic texture attack that requires a texture template image, we randomly choose one image from the training dataset as the template. We use adversarial weight of 500 and optimization step of 3 in a limited-memory BFGS optimizer as in \cite{semantic_attack}. For PGD attack, we use $\ell_\infty=8/255$ for adversarial training and $\ell_\infty=4/255$ during evaluation. For HSV attack, we set the random trail number to 100 as in \cite{hosseini2018semantic}. For feature space adversarial training, due to computational limit, we set the optimization step to 400 and adversarially train the model for 25 epochs. For other adversarial training, we train for 100 epochs. We use the Adam optimizer with a learning rate 1e-3, and set the batch size to 64 for all the adversarial training. A typical learning curve of FS adversarial training can be found at Figure~\ref{fig:learningcurve}.

During attack evaluation, we randomly sample 1000 images from the test set for untargeted attack. For the feature argumentation attack, we use the same parameters
as discussed in Evaluation section except for the number of attack steps, which is increased to 2000 during evaluation. It is a common practice when evaluating the performance of attacks.



Note that our settings are similar to those in pixel space adversarial training as we are not aware of existing adversarial training practice for feature space.


The results are presented in Table~\ref{tab:adv_train}.
The first column denotes the adversarially trained models. The second column shows the normal model accuracy. The following four columns present the attack success rate of various attacks for the different models. 
We observe that PGD adversarial training is most effective against PGD attack and has some effectiveness against HSV attack, but not FS or SM attack.
The semantic adversarial training can reduce the SM attack success rate by 34\% and the PDG attack success rate by 38\%, suggesting the two may share some common attack space. But it is not effective for FS or HSV attack. Feature space adversarial training can reduce the FS attack success rate by 27\% and the HSV attack by 13\%, but not others. 
This suggests that feature space attack aims at different attack space than PGD and semantic attacks.
Note that the improvement of feature space adversarial training is not as substantial as PGD training in the pixel space. We believe that it is because either our study is preliminary and more setups need to be explored; or, feature space adversarial training may be inherently harder and demand new methods. We will leave it to our future study.

We further study the effect of model capacities and optimization steps during adversarial training on the robustness of feature space hardened models as shown in Table~\ref{tab:adv_ablation}.
For adversarial training with optimization steps of 200 or more, we train the model for 25 epochs (instead of 100) due to the computational limit.
The first column denotes different model capacities used for adversarial training. The second column shows the number of optimization steps used for feature space attack during training. To test how the model capacity affects the result of adversarial training, we correspondingly increase the width and depth of the model by using Wide-ResNet18~\cite{DBLP:journals/corr/ZagoruykoK16} and ResNet28. We can observe that with the increase of optimization steps, the initial improvement of robustness is significant. When the steps increase to 200 and beyond, the benefit  becomes limited. By using a deeper model, the result can be further improved by 6\%, while a wider model does not lead to evident improvement.

\end{document}